\documentclass[runningheads,a4paper]{llncs}[2015/06/24]

\usepackage{cmap}
\usepackage[T1]{fontenc}

\usepackage{graphicx}

\usepackage[ngerman,english]{babel}
\addto\extrasenglish{\languageshorthands{ngerman}\useshorthands{"}}

\usepackage[%
rm={oldstyle=false,proportional=true},%
sf={oldstyle=false,proportional=true},%
tt={oldstyle=false,proportional=true,variable=true},%
qt=false%
]{cfr-lm}
\usepackage[math]{blindtext}

\usepackage{cite}

\usepackage{paralist}

\usepackage{csquotes}

\usepackage{microtype}

\usepackage{url}
\makeatletter
\g@addto@macro{\UrlBreaks}{\UrlOrds}
\makeatother

\usepackage{xcolor}

\usepackage{pdfcomment}
\hypersetup{hidelinks,
   colorlinks=true,
   allcolors=black,
   pdfstartview=Fit,
   breaklinks=true}
\usepackage[all]{hypcap}

\usepackage[capitalise,nameinlink]{cleveref}
\crefname{section}{Sect.}{Sect.}
\Crefname{section}{Section}{Sections}

\usepackage{xspace}

\DeclareFontFamily{U}{MnSymbolC}{}
\DeclareSymbolFont{MnSyC}{U}{MnSymbolC}{m}{n}
\DeclareFontShape{U}{MnSymbolC}{m}{n}{
    <-6>  MnSymbolC5
   <6-7>  MnSymbolC6
   <7-8>  MnSymbolC7
   <8-9>  MnSymbolC8
   <9-10> MnSymbolC9
  <10-12> MnSymbolC10
  <12->   MnSymbolC12%
}{}
\DeclareMathSymbol{\powerset}{\mathord}{MnSyC}{180}

\begin{document}

\title{Pediatric Bone Age Assessment Using Deep Convolutional Neural Networks}
\titlerunning{Deep Learning for Bone Age Assessment}


%
\author{Vladimir Iglovikov\inst{1}, Alexander Rakhlin\inst{2}, Alexandr A.Kalinin\inst{3},\\ \and Alexey Shvets\inst{4} }
\authorrunning{Iglovikov et al.}

\institute{
Lyft Inc., San Francisco, CA 94107, USA\\
\email{iglovikov@gmail.com} \and Neuromation OU, Tallinn, 10111 Estonia\\
\email{rakhlin@neuromation.io} \and University of Michigan, Ann Arbor, MI 48109, USA\\ \email{akalinin@umich.edu} \and Massachusetts Institute of Technology, Cambridge, MA 02142, USA\\
\email{shvets@mit.edu}
}
\maketitle

\begin{abstract}
Skeletal bone age assessment is a common clinical practice to diagnose endocrine and metabolic disorders in child development. In this paper, we describe a fully automated deep learning approach to the problem of bone age assessment using data from the 2017 Pediatric Bone Age Challenge organized by the Radiological Society of North America. The dataset for this competition consists of 12,600 radiological images. Each radiograph in this dataset is an image of a left hand labeled with bone age and sex of a patient. Our approach utilizes several deep neural network architectures trained end-to-end. We use images of whole hands as well as specific parts of a hand for both training and prediction. This approach allows us to measure the importance of specific hand bones for automated bone age analysis. We further evaluate the performance of the suggested method in the context of skeletal development stages. Our approach outperforms other common methods for bone age assessment.
\end{abstract}


\section{Introduction}\label{sec:intro}

\begin{figure}[!b]
\centering
\includegraphics[width=8cm]{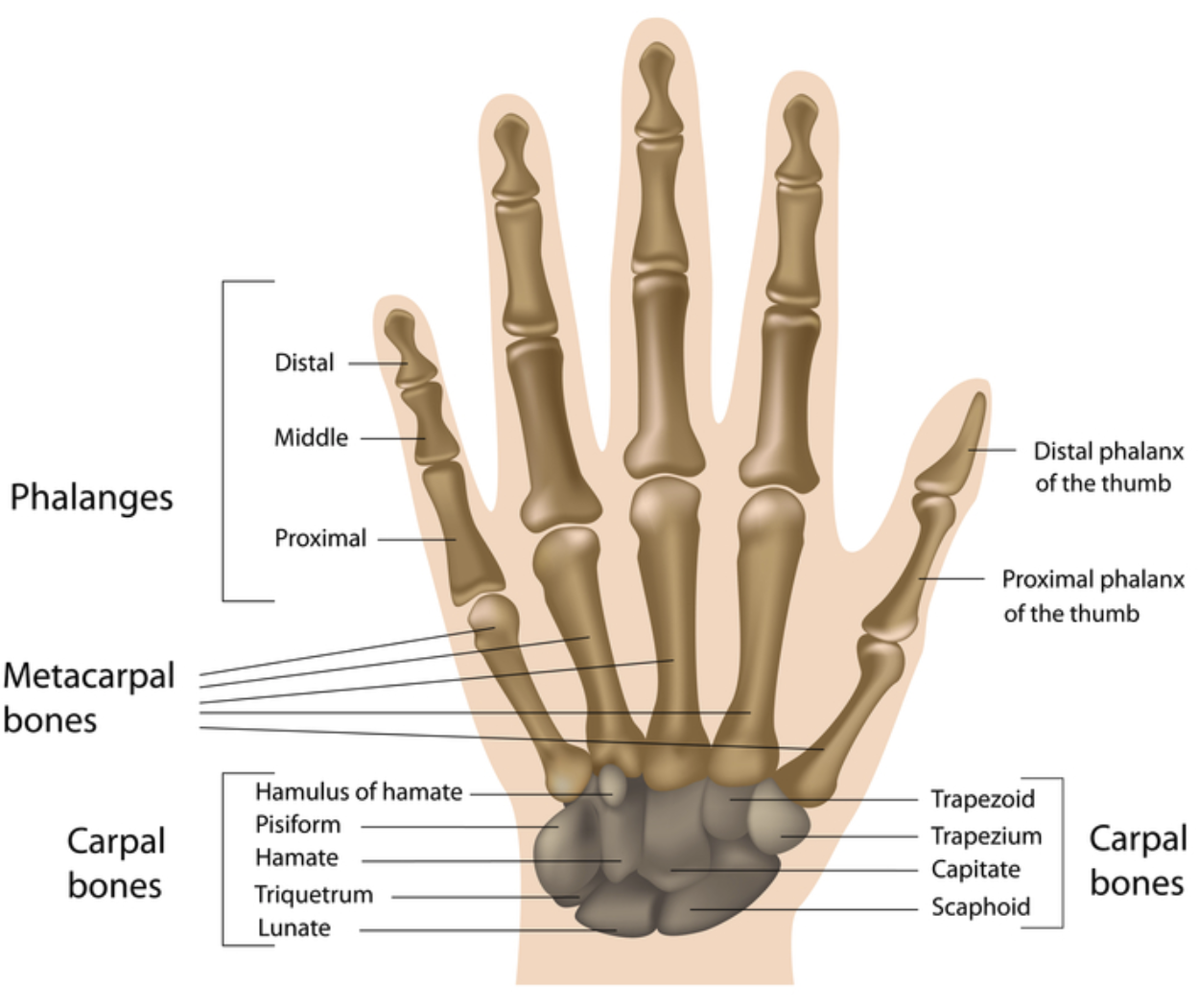}
\caption{Bones of a human hand and wrist (adapted from the Human Anatomy library \cite{human_anatomy}).}
\label{fig::hand_and_wrist}
\end{figure}

During organism development the bones of the skeleton change in size and shape, and thus a difference between a child's assigned bone and chronological ages might indicate a growth problem. Clinicians use bone age assessment in order to estimate the maturity of a child's skeletal system. Bone age assessment methods usually start with taking a single X-ray image of the left hand from the wrist to fingertips, see \cref{fig::hand_and_wrist}. Bones in the X-ray image are compared with radiographs in a standardized atlas of bone development. Such bone age atlases are based on large numbers of radiographs collected from children of the same sex and age.

Over the past decades, the bone age assessment procedure has been performed manually using either the Greulich and Pyle (GP) \cite{greulich1959radiographic} or Tanner-Whitehouse (TW2) \cite{tanner1983assessment} methods. The GP procedure determines the bone age by comparing the patient's radiograph with an atlas of representative ages. The TW2 technique is based on a scoring system that examines 20 specific bones. In both cases, bone assessment procedure requires a considerable time. Only recently software solutions, such as BoneXpert \cite{thodberg2009bonexpert}, have been developed and approved for the clinical use in Europe. BoneXpert uses the Active Appearance Model (AAM) \cite{cootes1992training}, a computer vision algorithm, which reconstructs the contours of 13 bones of a hand. Then the system determines the overall bone age according to their shape, texture, and intensity based on the GP or TW techniques.  However, it is sensitive to the image quality and does not utilize the carpal bones, despite their importance for skeletal maturity assessment in infants and toddlers.

Recently, deep learning-based approaches demonstrated performance improvements over conventional machine learning methods for many problems in biomedicine \cite{ching2017opportunities, kalinin2018deep}. In the domain of medical imaging, convolutional neural networks (CNN) have been successfully used, for example, for diabetic retinopathy screening \cite{rakhlin2017diabetic}, for breast cancer histology image analysis \cite{rakhlin2018deep}, bone disease prediction \cite{tiulpin2018automatic}, and other problems \cite{ching2017opportunities}. In the case of bone age assessment, a manually performed procedure requires around 30 minutes of doctor's time per a patient. When the same procedure is done using software based on classical computer vision methods, it takes 1-5 minutes, but still requires substantial doctoral supervision and expertise. Deep learning based methods allow to avoid feature engineering by automatically learning the hierarchy of discriminative features directly from a set of labeled examples. Using a deep learning approach processing of one image typically takes less than 1 sec, while accuracy of these methods in many cases exceeds that of conventional methods. Deep neural network based solutions for bone age assessment from hand radiographs were suggested before \cite{lee2017fully, spampinato2017deep, larson2017performance}. However, these studies did not perform a numerical evaluation of the performance of their models using different hand bones. In addition, we find that the performance of deep learning models for bone age assessment can be further improved with better preprocessing and training networks on radiographs from scratch instead of fine-tuning from natural image domain.

\begin{figure}[!t]
\centering
\includegraphics[width=10cm]{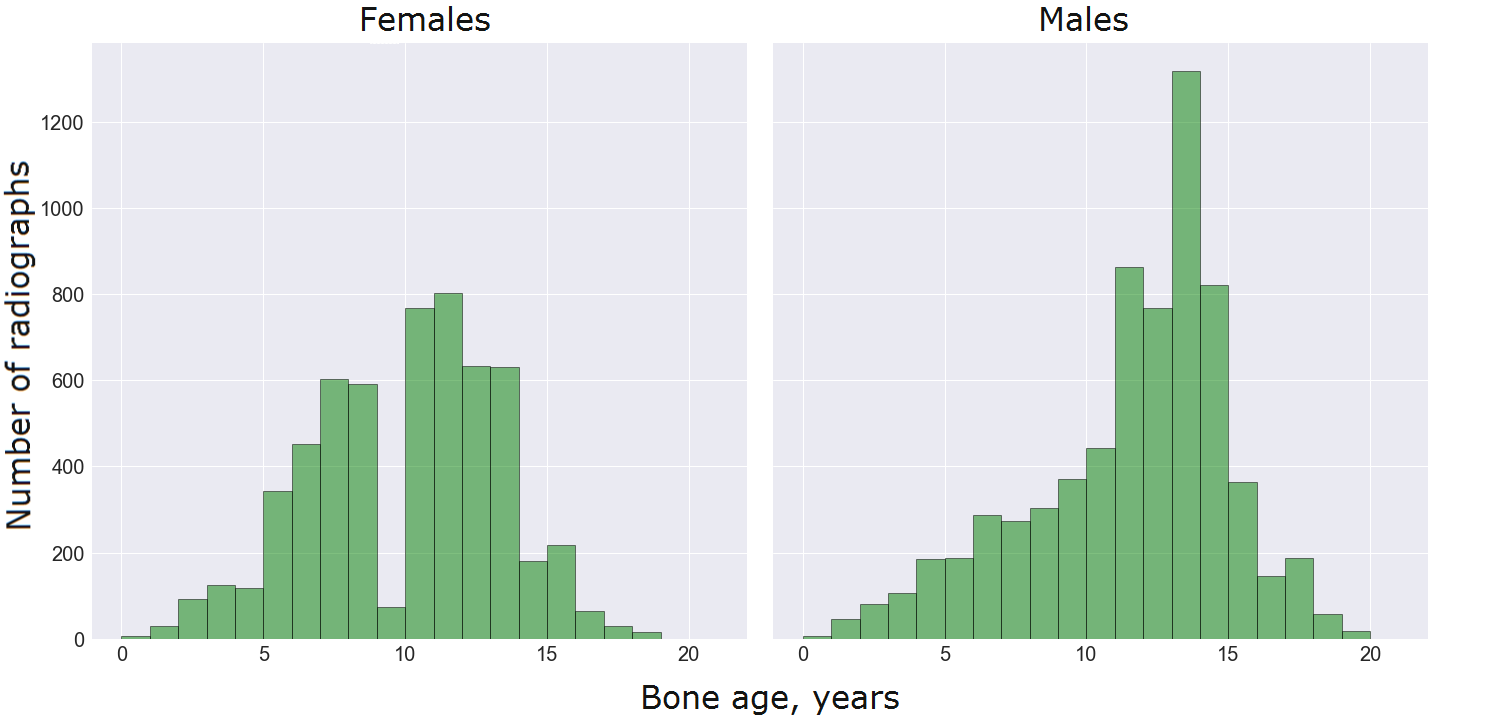}
\caption{Bone age distribution for females and males in the training dataset.}
\label{fig::male_femail_dist}
\end{figure}

In this paper, we present a novel deep learning based method for bone age assessment. We validate the performance of this method using the data from the 2017 Pediatric Bone Age Challenge organized by the Radiological Society of North America (RSNA) \cite{rsna2017challenge}. This data set is now freely available and can be accessed at \cite{rsna2017dataset}. In our approach, we first preprocess radiographs by segmenting the hand, normalizing contrast, detecting key points and using them to register segmented hand images. Then, we train several deep network architectures using different parts of images to evaluate how various bones contribute to the models' performance across four major skeletal development stages. Finally, we compare predictions across different models and evaluate the overall performance of our approach. We demonstrate that the suggested method is more robust and demonstrates superior performance compared to other commonly used solutions.

\section{Preprocessing}
The goal of the first step in the preprocessing pipeline is to extract a region of interest (a hand mask) from the image and remove all extraneous objects. Images were collected at different hospitals and simple background removal methods did not produce satisfactory results. Thus, there is a compelling need for a reliable hand segmentation technique. However, this type of algorithms typically requires large manually labeled training set. To alleviate labeling costs, we employ a technique called positive mining. In general, positive mining is an iterative procedure where manual labeling is combined with automatic processing. It allows us quickly obtain accurate masks for all images in the training set.

Overall our preprocessing method includes binary image segmentation as a first step and then the analysis of connected components for the post-processing of segmentation results. For the image segmentation, we use U-Net deep network architecture originally proposed in \cite{ronneberger2015u}. U-Net is capable of learning from a relatively small training set that makes it a good architecture to combine with positive mining. In general, the U-Net architecture consists of a contracting path to capture context and a symmetric expanding path that enables precise localization. The contracting path follows the typical architecture of a convolutional network with convolution and pooling operations and progressively downsampled feature maps. Every step in the expansive path consists of an upsampling of the feature map followed by a convolution. Hence, the expansive branch increases the resolution of the output. In order to localize, upsampled features in the expansive path are combined with the high resolution features from the contracting path via skip-connections \cite{ronneberger2015u}. This architecture proved itself very useful for segmentation problems with limited amounts of data, e.g. see \cite{iglovikov2017satellite}. We also employ batch normalization technique to improve convergence during training \cite{ioffe2015batch}.

In our algorithms, we use the generalized loss function
\begin{equation}
\label{free_en}
L=H-\log J\,,
\end{equation}
where $H$ is a binary cross entropy that defined as 
\begin{equation}
H=-\frac{1}{n}\sum\limits_{i=1}^n(y_i\log \hat{y}_i+(1-y_i)\log (1-\hat{y}_i))\,,
\end{equation}
where $y_i$ is a binary value of the corresponding pixel $i$ ($\hat{y}_i$ is a predicted probability for the pixel). In the second term of \cref{free_en}, $J$ is a differentiable generalization of the Jaccard Index
\begin{equation}
J=\frac{1}{n}\sum\limits_{i=1}^n\left(\frac{y_i\hat{y}_i}{y_{i}+\hat{y}_i-y_i\hat{y}_i}\right)\,.
\end{equation}
For more details see \cite{iglovikov2017satellite}.

\begin{figure}[!t]
\centering
\includegraphics[width=10cm]{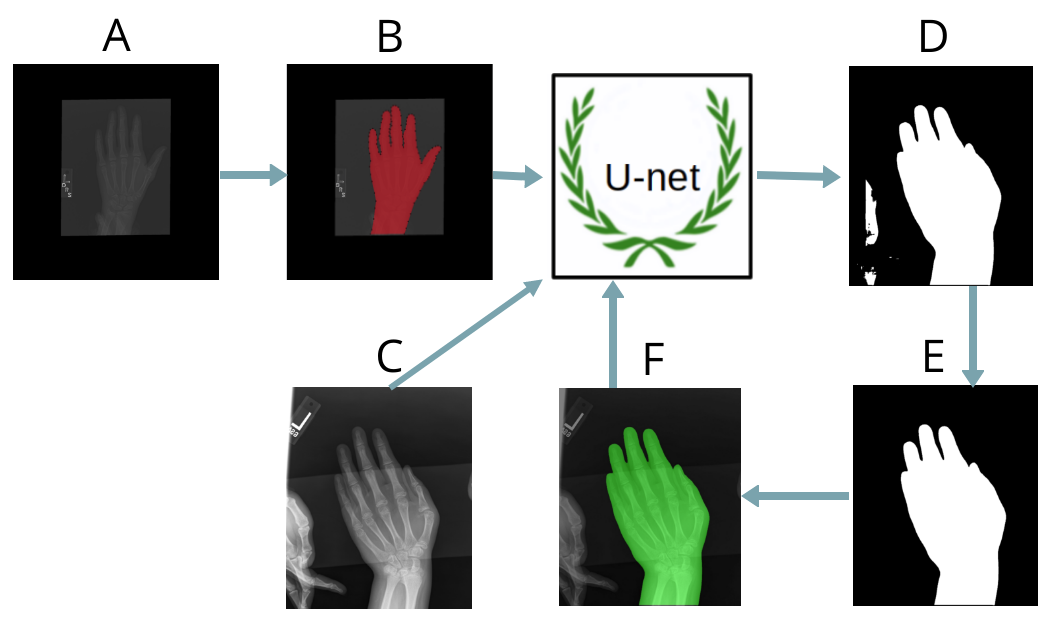}
\caption{Iterative procedure of positive mining utilizing U-Net architecture for image segmentation: (A) raw input data; (B) mask manually labeled with the online annotation tool Supervisely \cite{supervisely}; (C) new data; (D) raw prediction; (E) post processed prediction; (F) raw image with mask plotted together for visual inspection.}
\label{fig::unet_procedure}
\end{figure}

In this work, we first manually label 100 masks using the online annotation service Supervisely \cite{supervisely} that takes approximately 2 min to process each image. These masks are used to train the U-Net model that then is used to perform hand segmentation on the rest of the training set. Since each image is supposed to contain only one hand, for each prediction we remove all connected components except for the largest one that is kept and further subjected to the standard hole filling protocol. This procedure allows us to predict hand masks in the unlabeled train set, however, visual inspection reveals that the quality of mask predictions is inconsistent and requires additional improvements. Thus, we visually inspect all predicted masks and keep only those of acceptable quality, while discarding the rest. This manual process allows us to curate approximately 3-5 images per second. Expanding the initial training set with the additional good quality masks increases the size of the labeled images for segmentation procedure and improves segmentation results. To achieve an acceptable quality on the whole training set, we repeat this procedure 6 times. Finally, we manually label approximately 100 of the corner cases that U-Net is not able to capture well. The whole iterative procedure is schematically shown in \cref{fig::unet_procedure}

Original GP and TW2 methods focus on specific hand bones, including phalanges, metacarpal and carpal bones, see \cref{fig::hand_and_wrist}. So, we choose to train separate models on several specific regions in high resolution and subsequently evaluate their performance. To correctly locate these regions it is necessary to transform all the images to the same size and position, i.e. to register them in one coordinate space. Hence, our model comprises two sub-models: image registration and bone age assessment of a specific region.

\section{Key points detection}

\begin{figure}[!b]
\centering
\includegraphics[width=8cm]{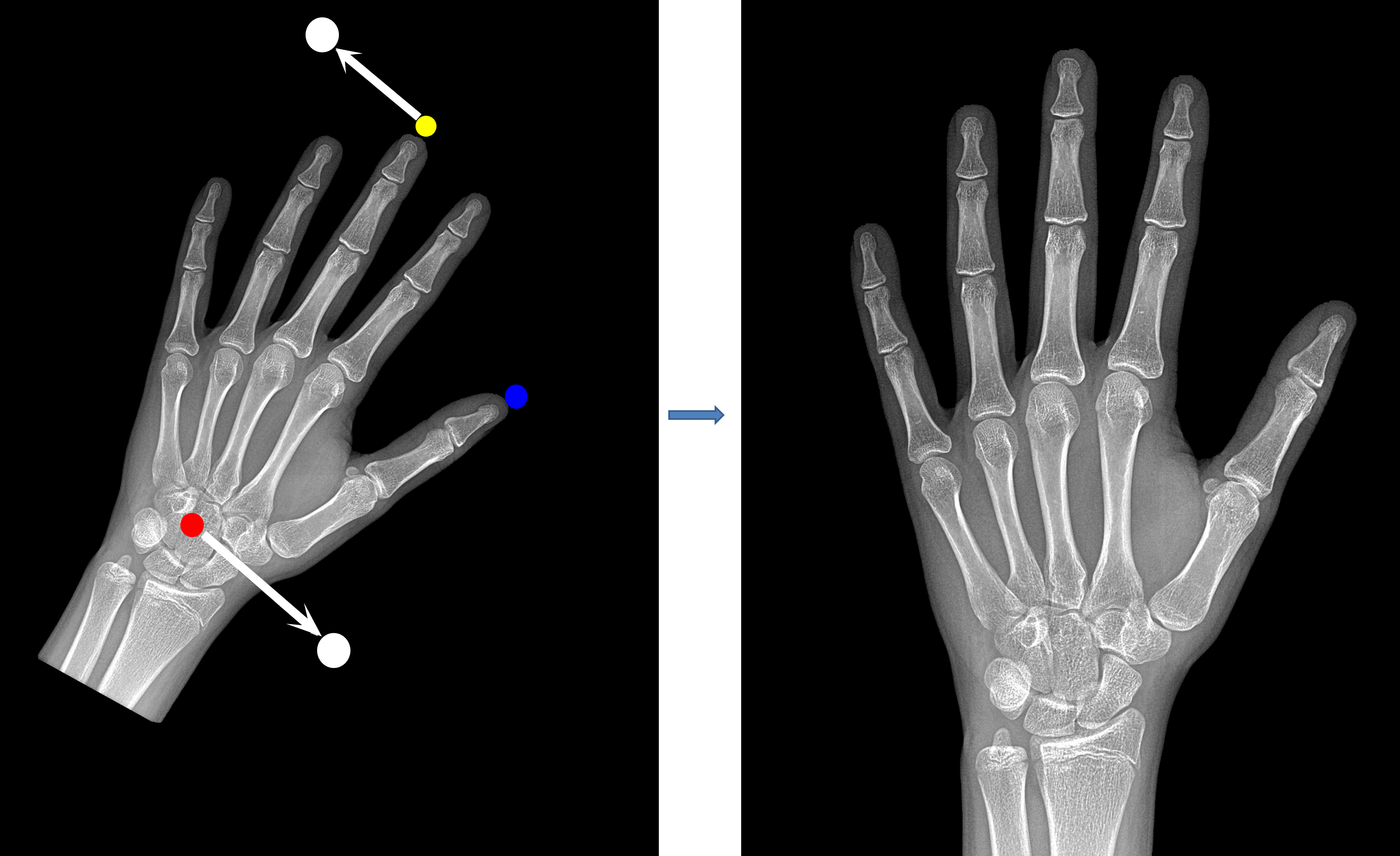}
\caption{Image registration. (Left) Key points: the tip of the middle finger (the yellow dot), the center of the capitate (the red dot), the tip of the thumb (the blue dot). Registration positions: for the tip of the middle finger and for the center of the capitate (white dots). (Right) Registered image after key points are found and the affine transformation and scaling are applied.}
\label{fig::keypoints}
\end{figure}

One of our goals is to evaluate the importance of specific regions of a hand for the automated bone age assessment. This opens remarkable opportunity of running a model on smaller image crops with higher resolution that might result in reduced processing time and higher accuracy. To crop a specific region, we have to \emph{register} hand radiographs, or in other words, align them into a common coordinate space. To this end, we first detect coordinates of several specific key points of a hand. Then, we calculate affine transformation parameters (zoom, rotation, translation, and mirror) to fit the image into the desired position (\cref{fig::keypoints}).

Three characteristic points on the image are chosen: the tip of the distal phalanx of the third finger, tip of the distal phalanx of the thumb, and center of the capitate. All images are re-scaled to the same resolution: $2080\times1600$ pixels, and padded with zeros when necessary. To create training set for key points model, we manually label 800 radiographs. Pixel coordinates of key points serve as training targets for our regression model. Registration procedure is shown in \cref{fig::keypoints}.

\begin{figure}[!t]
\centering
\includegraphics[width=12cm]{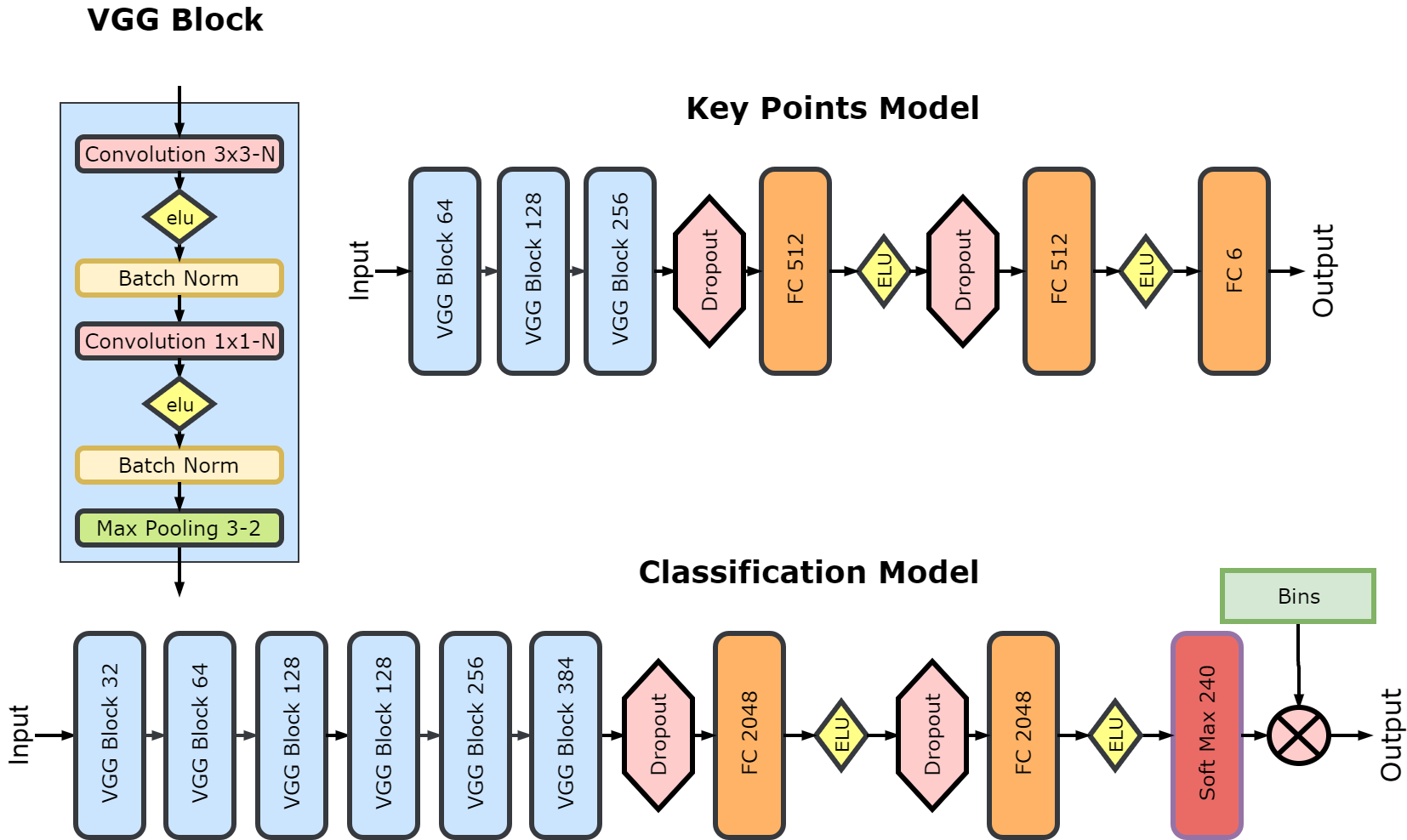}
\caption{VGG-style neural network architectures for regression (top) and classification (bottom) tasks.}
\label{fig::vgg}
\end{figure}

Key points model is implemented as a deep convolutional neural network, inspired by a popular VGG family of models \cite{simonyan2014vgg}, with a regression output. The network architecture is schematically shown in \cref{fig::vgg}. The VGG module consists of two convolutional layers with the Exponential Linear Unit (ELU) activation function \cite{clevert2015elu}, batch normalization \cite{ioffe2015batch}, and max pooling. The input image is passed through a stack of three VGG blocks followed by three Fully Connected layers. VGG blocks consist of 64, 128, 256 convolution layers respectively. For better generalization, dropout units are applied in-between. The model is trained with Mean Squared Error loss function (MSE) using Adam optimizer \cite{kingma2014adam}:

\begin{equation}
MSE=\frac{1}{n}\sum\limits_{i=1}^n(\hat{y}_i-y_i)^2\,.
\end{equation}

To reduce computational costs, we downscale input images to 130x100 pixels. At the same time, target coordinates for key points are re-scaled from $[0, 2079]\times[0, 1599]$ to the uniform square $[-1, 1]\times[-1, 1]$. At the inference stage after the model detects key points, we project their coordinates back to the original image size, i.e $2080\times1600$ pixels. To improve generalization of our model, we apply input augmentations: rotation, translation and zoom. The model output consists of 6 coordinates, 2 for every key point.

\begin{figure}[!t]
\centering
\includegraphics[width=12cm]{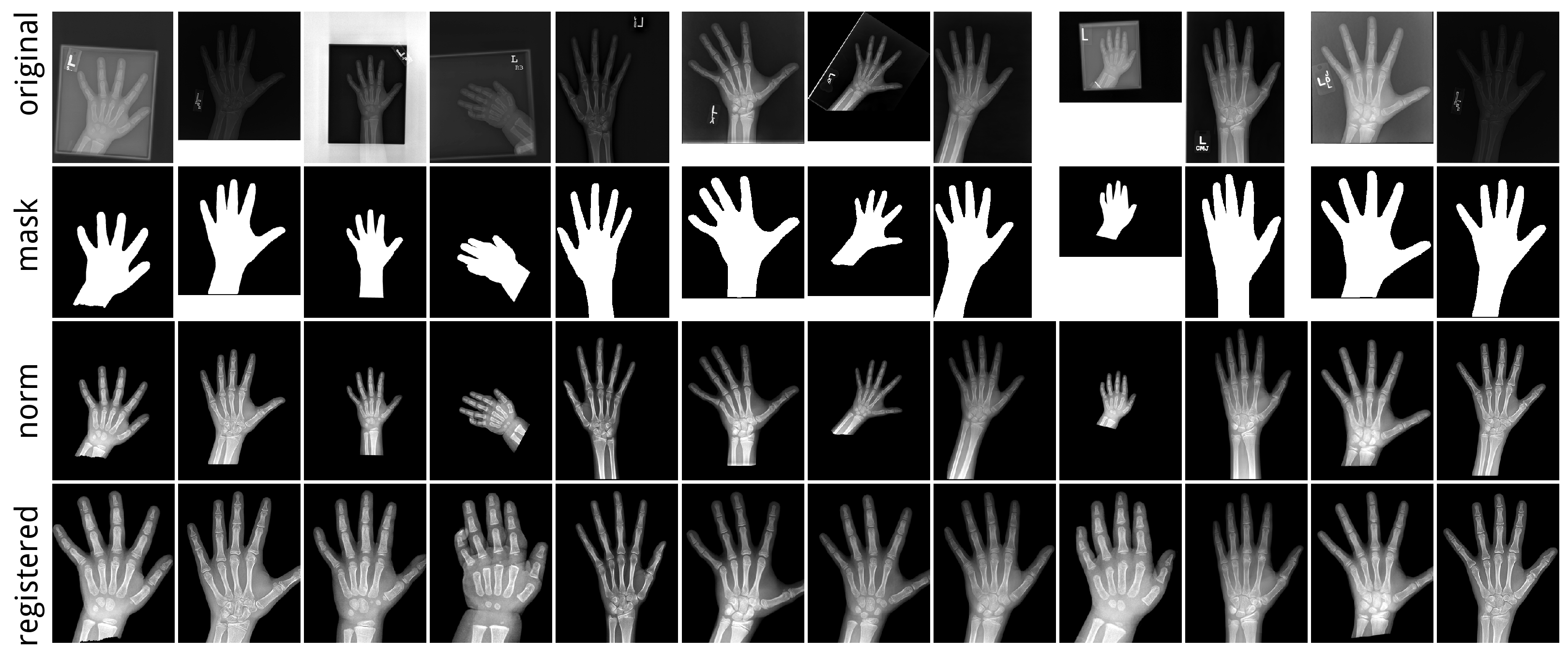}
\caption{Preprocessing pipeline: (first row) original images; (second row) binary hand masks that are applied to the original images to remove background; (third row) masked and normalized images; (bottom row) registered images.}
\label{fig::grid_large}
\end{figure}

At the next step, we calculate affine transformations (zoom, rotation, translation) for all radiographs. Our goal is to preserve proportions of an image and to fit it into uniform position such that for every image: 1) the tip of the middle finger is aligned horizontally and positioned approximately 100 pixels below the top edge of the image; 2) the capitate (see \cref{fig::hand_and_wrist}) is aligned horizontally and positioned approximately 480 pixels above the bottom edge of the image. By convention, bone age assessment uses radiographs of the left hand, but sometimes the images in the dataset get mirrored. To detect these images and adjust them appropriately the key point for the thumb is used. The results of the segmentation, normalization and registration are shown in the fourth row of \cref{fig::grid_large}.

\section{Bone age assessment model}

Although CNNs are more commonly used in classification tasks, bone age assessment is a regression task by nature. In order to access performance in both settings, we compare two types of CNNs: regression and classification. Both models share similar architectures and training protocols, and only differ in two final layers.

\subsection{Regression model}
Our first model is a VGG-style CNN \cite{simonyan2014vgg} with a regression output. This network represents a stack of six convolutional blocks with 32, 64, 128, 128, 256, 384 filters followed by two fully connected layers of 2048 neurons each and a single output (see \cref{fig::vgg}). The input size varies depending on the considered region of an image, \cref{fig::ZonesABC}. For better generalization, we apply dropout layers before the fully connected layers. For regression targets, we scale bone age in the range $[-1, 1]$. The network is trained by minimizing Mean Absolute Error (MAE):
\begin{equation}
MAE=\frac{1}{n}\sum\limits_{i=1}^n|\hat{y}_i-y_i|    
\end{equation}
with Adam optimizer. We begin training with the learning rate $10^{-3}$ and then progressively lower it to $10^{-5}$. Due to a limited data set size, we use train time augmentation with zoom, rotation and shift to avoid overfitting.

\subsection{Classification model}
The classification model (\cref{fig::vgg}) is similar to the regression one, except for the two final layers. First, we assign each bone age a class. Bone ages expressed in months, hence, we assume 240 classes overall. The second to the last layer is a softmax layer with 240 outputs. This layer outputs vector of probabilities of 240 classes. The probability of a class takes a real value in the range $[0, 1]$. In the final layer, the softmax layer is multiplied by a vector of distinct bone ages uniformly distributed over 240 integer values $[0, 1, ..., 238, 239]$. Thereby, the model outputs single value that corresponds to the expectation of the bone age. We train this model using the same protocol as the regression model.

\subsection{Region-specific modelling}
\begin{figure}[!b]
\centering
\includegraphics[width=5cm]{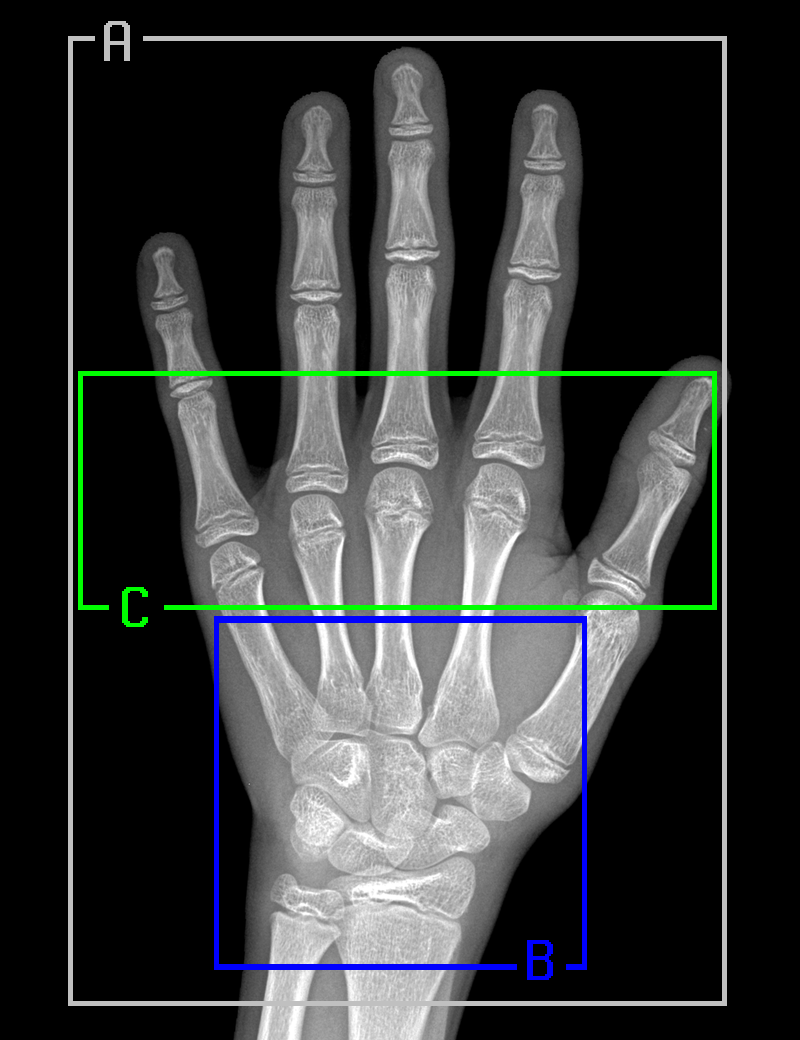}
\caption{A registered radiograph with three specific regions: (A) a whole hand; (B) carpal bones; (C) metacarpals and proximal phalanges.}
\label{fig::ZonesABC}
\end{figure}

In accordance with the targeted features of skeletal development stages described in \cite{greulich1959radiographic, tanner1983assessment, gilsanz2005hand}, we crop three specific regions from registered radiographs ($2080\times1600$ pixel), as shown in \cref{fig::ZonesABC}:

\begin{enumerate}
\item whole hand ($2000\times1500$ pixel)\\
\item carpal bones ($750\times750$ pixel)\\
\item metacarpals and proximal phalanges ($600\times1400$ pixel)
\end{enumerate}

\subsection{Experiment setup}
We split labeled radiographs into two sets preserving sex ratio. The training set contains 11,600 images, and validation set contains 1,000 images. We create several models with a breakdown by:

\begin{enumerate}
\item type (regression, classification)\\
\item sex (males, females, mixed)\\
\item region (A, B, C)
\end{enumerate}

Given these conditions, we produce 18 basic models ($2\times3\times3$). Furthermore, we construct several meta-models as a linear average of regional models and, finally, an average of different models.

\section{Results}

\begin{figure}[!b]
\centering
\includegraphics[width=12cm]{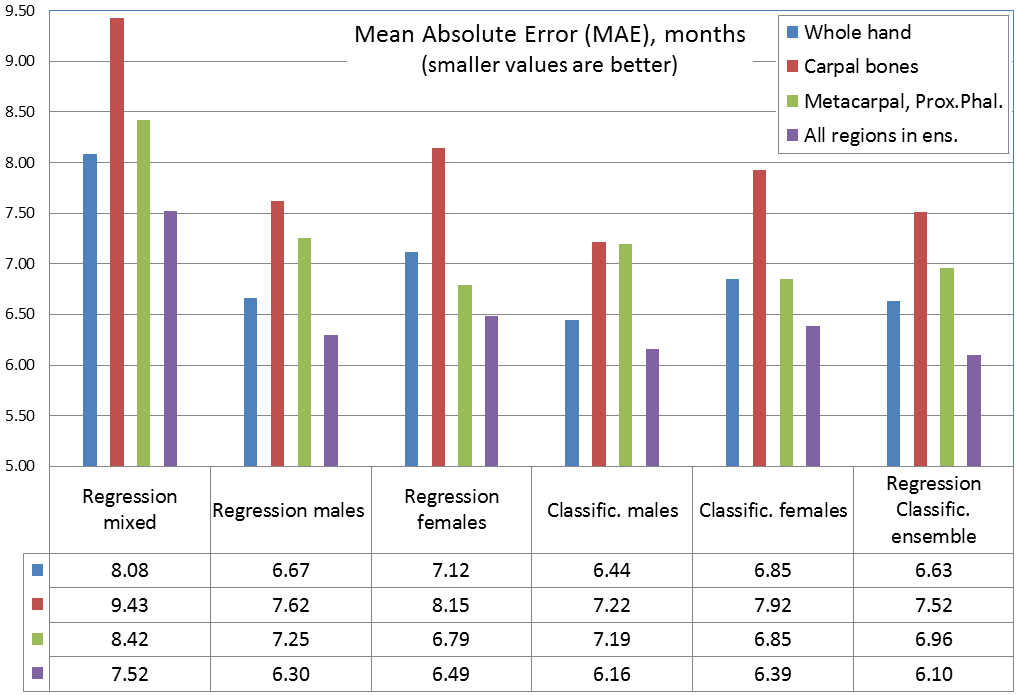}
\caption{Mean absolute errors on the validation data set for regression and classification models for different bones and sexes. Colors correspond to different regions. Table: regions are shown in rows, models in columns. There is a total of 15 individual models and 9 ensembles.}
\label{fig::diagr_en}
\end{figure}

The performance of all models is evaluated on validation data set, as presented in \cref{fig::diagr_en}. The leftmost column represents the performance of a regression model for both sexes.
The region of metacarpals and proximal phalanges (region C) has higher accuracy with MAE equal to 8.42 months.
MAE of the whole image (region A) is 8.08 months. The linear ensemble of the three regional models outperforms all of the above models with MAE 7.52 months (bottom row). This regional pattern MAE(B) $>$ MAE(C) $>$ MAE(A) $>$ MAE (ensemble) is further observed for other model types and patient cohorts with few exceptions. Separate regression models for male and female cohorts (second and third columns) demonstrated higher accuracy when compared to those trained on a mixed population. The ensemble of regional models has MAE equal to 6.30 months for males and 6.49 months for females (bottom row). For males, the top performing region is the whole hand (A) that has MAE equal to 6.67 months. In contrast, for the female cohort region of metacarpals and proximal phalanges (C) has MAE equal to 6.79 months and this result is the most accurate across the three regions. Classification models (fourth and fifth columns) perform slightly better than regression networks. The ensemble of the regional models has MAE equal to 6.16 months for males and 6.39 months for females (bottom row). For the male cohort the whole hand region (A) has the highest accuracy with MAE equal to 6.44 months. For the female cohort the result produced using the metacarpals and proximal phalanges region (C) is on par with that obtained using the whole hand and has MAE equal to 6.85 months for both of them.

In the last column, we analyze the ensemble of classification and regression models. As shown, the MAEs of regional models follow overall pattern MAE(B) $>$ MAE(C) $>$ MAE(A). The ensemble of the regional models (bottom row) has the best accuracy with MAE equal to 6.10 months. This result outperforms state-of-the-art results of the BoneXpert software (SD 0.72 years\footnote{http://www.bonexpert.com/products/the-bonexpert-product}) and the recent application of deep neural networks (RMSE 0.82-0.93 years) \cite{lee2017fully}.

\begin{figure}[t!]
\centering
\includegraphics[width=12cm]{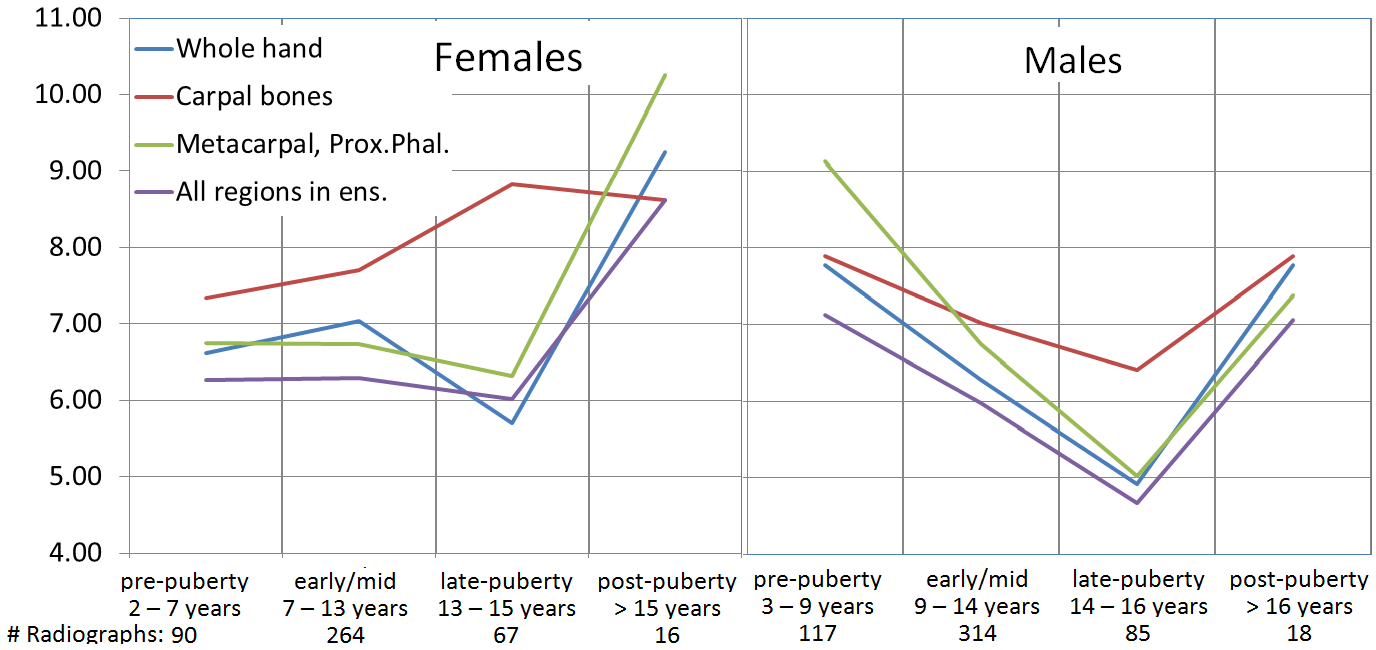}
\caption{Mean absolute error in months as a function of skeletal development stages for different sexes. Different colors on the plot correspond to different regions of a radiograph. For males and females the development stages are labelled at the bottom of each plot.}
\label{fig::stages}
\end{figure}

Next we evaluate our method in the context of the age distribution. Following \cite{lee2017fully, gilsanz2005hand}, we consider four major skeletal development stages: pre-puberty, early-and-mid puberty, late puberty, and post-puberty. Infant and toddler categories were excluded due to scarcity of the data: the development data set contained only 90 radiographs for bone age less than 24 months and only 9 of them were presented in our validation subset. The model accuracies for the four skeletal development stages, for different regions and sexes are depicted in the \cref{fig::stages}. Note that the skeletal development stages are different for males and females. Based on this data, we report two important findings. 

First, unlike Lee \textit{et al.} \cite{lee2017fully}, we do not observe better results when training on carpal bones compared to other areas. With the two exceptions, the metacarpals and proximal phalanges provide better accuracy than the carpal bones do. These exceptions are pre-puberty for the male cohort and post-puberty for the female cohort, where the accuracy of the carpal bones is higher. However, the validation dataset sizes for these two skeletal development stages (117 for males in pre-puberty and 16 for females in post-puberty) are too small to draw statistically significant conclusion. At the same time, we notice that Gilsanz and Ratib \cite{gilsanz2005hand} proposed carpal bones as the best predictor of skeletal maturity only in infants and toddlers. Thereafter, we find no sound evidence to support the suggestion that the carpal bones can be considered as the best predictor in the pre-puberty, see \cite{lee2017fully}.

The second interesting finding is the influence of the dataset on the accuracy of a model. For the both sexes the accuracy peaks at the late-puberty, the most frequent age in the data set. This dependency is particularly evident in the male cohort, where the accuracy essentially mirrors data distribution (see \cref{fig::male_femail_dist}). This finding is very important as it suggests a straightforward way for the future improvement of the method.

\section{Conclusion}
In this study, we investigate the application of deep convolutional neural networks to the problem of the automatic bone age assessment. The automatic bone age assessment system based on our approach can estimate skeletal maturity with accuracy similar to that of an expert radiologist and surpasses existing automated models, e.g. see \cite{lee2017fully}. In addition, we numerically evaluate different zones of a hand in bone age assessment. We find that bone age assessment could be done just for carpal bones or for metacarpals and proximal phalanges with around 10-15$\%$ increase in error compared to the whole hand assessment. Therefore, we can establish bone age using just part of the radiogram with high enough quality, lowering computational overheads.

Despite the challenging quality of the radiographs, our approach succeeds in image preprocessing, cleaning and standardization. These transformations, in turn, greatly help in improving the robustness and performance of deep learning models. Moreover, the accuracy of our approach can be improved even further. First, our solution could be easily combined with other, more complex network architectures, such as Resnet \cite{he2016resnet}. Another way to improve it is to substantially extend the training data set by additional examples. Furthermore, bone ages that serve as labels for training may also be refined based on work of independent experts. Thereby, implementing these simple steps could potentially lead to a development of a state-of-the-art bone assessment software system. It would have a potential for the deployment in the clinical environment in order to help doctors in making a final age bone assessment decision accurately and in real time, with just one click. Moreover, a cloud-based system could be deployed for this problem and process radoigraphs independently of their origin. This potentially could help thousands of doctors to get a qualified evaluation even in hard-to-reach areas.


To conclude, in the final stage of the RSNA2017 Pediatric Bone Age Assessement challenge our solution has been evaluated by organizers using the test set. This data set consisted of 200 radiographs equally divided between sexes. All labels have been hidden from participants. Based on organizers' report our method achieved MAE equal to 4.97 months, which is higher compared to the performance on the data withheld from the training set. The explanation of such improvement may be hidden in more accurate labelling of the test set or better quality of the radiographs. Each of radiographs in the test data set was evaluated and crosschecked by three experts independently, compared to a one expert that labelled training radiographs. This again demonstrate the importance of the input from domain experts.


\subsubsection*{Acknowledgments}
The authors would like to thank Open Data Science community \cite{ods} for many valuable discussions and educational help in the growing field of machine/deep learning.

\bibliographystyle{splncs03}
\bibliography{paper}


\end{document}